\documentclass[runningheads]{llncs}
\usepackage{booktabs}
\usepackage{multirow}
\usepackage{hyperref}
\usepackage{url}
\usepackage{graphicx} % DO NOT CHANGE THIS
\usepackage{caption}
\usepackage{amsmath}        % 数学公式、align、equation等
\usepackage{amssymb}        % 数学符号 (\mathbb, \mathcal等)
\usepackage[T1]{fontenc}
\usepackage{graphicx,verbatim}
\begin{document}
\title{KEPIL: Knowledge-Enhanced Prompt-Image Learning for Prompt-Robust Disease Detection}
%\titlerunning{Abbreviated paper title}
% If the paper title is too long for the running head, you can set
% an abbreviated paper title here
%
\begin{comment}  %% Removed for anonymized MICCAI submission
\author{First Author\inst{1}\orcidID{0000-1111-2222-3333} \and
Second Author\inst{2,3}\orcidID{1111-2222-3333-4444} \and
Third Author\inst{3}\orcidID{2222--3333-4444-5555}}
%
\authorrunning{F. Author et al.}
% First names are abbreviated in the running head.
% If there are more than two authors, 'et al.' is used.
%
\institute{Princeton University, Princeton NJ 08544, USA \and
Springer Heidelberg, Tiergartenstr. 17, 69121 Heidelberg, Germany
\email{lncs@springer.com}\\
\url{http://www.springer.com/gp/computer-science/lncs} \and
ABC Institute, Rupert-Karls-University Heidelberg, Heidelberg, Germany\\
\email{\{abc,lncs\}@uni-heidelberg.de}}

\end{comment}

\author{%
  Haozhe Luo\inst{1,3}
  \and Shelley Zixin Shu\inst{1}
  \and Ziyu Zhou\inst{2}
  \and Robert Berke\inst{3}
  \and Mauricio Reyes\inst{1,4}
}
\authorrunning{H.\ Luo et al.}
\institute{{%
ARTORG Center for Biomedical Engineering Research, University of Bern\\
\email{\{haozhe.luo,mauricio.reyes3\}@unibe.ch}}    
  \and
 Shanghai Jiao Tong University\\
  \and
 Kaiko.AI, Zurich, Switzerland \\
 \and
 Department of Radiation Oncology, Inselspital, Bern University Hospital and
University of Bern, Bern, Switzerland
}

\maketitle              % typeset the header of the contribution
\begin{abstract}
Vision--language models (VLMs) show promise for clinical decision support in radiology because they enable joint reasoning over radiological images and clinical text, thereby leveraging complementary clinical information. However, radiological findings are long-tailed in practice, leaving some conditions underrepresented and making zero-shot inference essential. Yet current CLIP-style medical VLMs are sensitive to prompt variations and often lack trustworthy external knowledge at inference time, which hinders reliable clinical deployment. We present \textit{KEPIL}, a prompt-robust framework that integrates curated medical knowledge to stabilize zero-shot generalization. KEPIL comprises: (i) \emph{dynamic prompt enrichment} using ontologies with LLM assistance, (ii) a \emph{semantic-aware contrastive loss} aligning embeddings of equivalent prompt variants via a dual-embedding objective, and (iii) \emph{entity-centric report standardization} to yield ontology-aligned representations. Across seven benchmarks, KEPIL achieves state-of-the-art zero-shot inference performance; under prompt-variation tests, it improves AUC by \(6.37\%\) on \textit{CheXpert} and by \(4.11\%\) on average. These results suggest that structured knowledge and robust prompt design are key to clinically reliable radiology-facing VLMs. Code will be released at https://github.com/Roypic/KEPIL.
% Authors must provide keywords and are not allowed to remove this Keyword section.

\end{abstract}
\section{Introduction}

Vision–language models (VLMs) have become a compelling paradigm for medical AI by enabling joint reasoning over radiological images and accompanying clinical text. Nevertheless, radiological findings exhibit pronounced long-tailed distributions and impose substantial expert-annotation burdens, leaving many conditions underrepresented during training \cite{zhang2024challenges,holste2022long,lin2025cxr}. In this context, zero-shot inference is indispensable: for rare diseases, available cases are too limited to support supervised learning. Notably, standardized medical ontologies and radiographic descriptors are often available even for underrepresented conditions, enabling their semantic alignment with well-characterized common diseases. Such knowledge-driven alignment facilitates reliable diagnosis in data-sparse regimes ~\cite{zhang2023knowledge,lin2025cxr,rahman2024xdt,zhang2025medunifier}, thereby advancing the practical readiness of VLMs for clinical deployment. Despite this potential, their practical deployment remains limited by two key challenges. First, current CLIP-style medical VLMs ~\cite{zhang2022contrastive,huang2021gloria,luo2025interplay,boecking2022making,wu2023medklip} are highly sensitive to variations in textual prompts. Most existing approaches rely on fixed or template-based prompts during training and inference ~\cite{zhang2023knowledge,wu2023medklip,lai2024carzero}, yet even minor changes in phrasing at test time can lead to substantial performance degradation as illustrated in Figure \ref{fig:compare_robustness}. This lack of robustness undermines their reliability in real-world clinical settings, where user inputs are naturally diverse. Second, robust generalization to rare or unseen diseases hinges on dependable external knowledge bases. However, prior work often constructs disease knowledge solely via LLMs, inviting hallucination and inconsistency ~\cite{huang2025survey,zhang2025siren}. \\
To address these issues, we propose \textit{KEPIL} (Knowledge-Enhanced Prompt Image Learning), a novel framework designed to improve both robustness to prompt variations and transparent zero-shot generalization.  We instead ground knowledge in curated resources (e.g., Radiopaedia, UMLS) and use LLMs only to draft and normalize text under ontology constraints. This ontology-aligned, verifiable knowledge base mitigates hallucination and stabilizes zero-shot inference beyond the training distribution. KEPIL incorporates structured medical knowledge through three key components: (1) dynamic prompt enrichment using medical ontologies and large language models to generate diverse, clinically grounded descriptions; (2) a semantic-aware contrastive loss that encourages consistency among embeddings of prompt variants while separating semantically different prompts; and (3) entity-centric report standardization that reduces sensitivity to linguistic variation. Extensive experiments on seven benchmark datasets demonstrate that KEPIL achieves state-of-the-art performance in zero-shot classification and segmentation, with notable improvements in robustness to prompt variation and enhanced recognition of rare conditions. These findings underscore the critical role of robustness to prompt in developing clinically reliable VLMs.

\section{Method}

\begin{figure*}[t!]
    \centering
    \includegraphics[width=11cm]{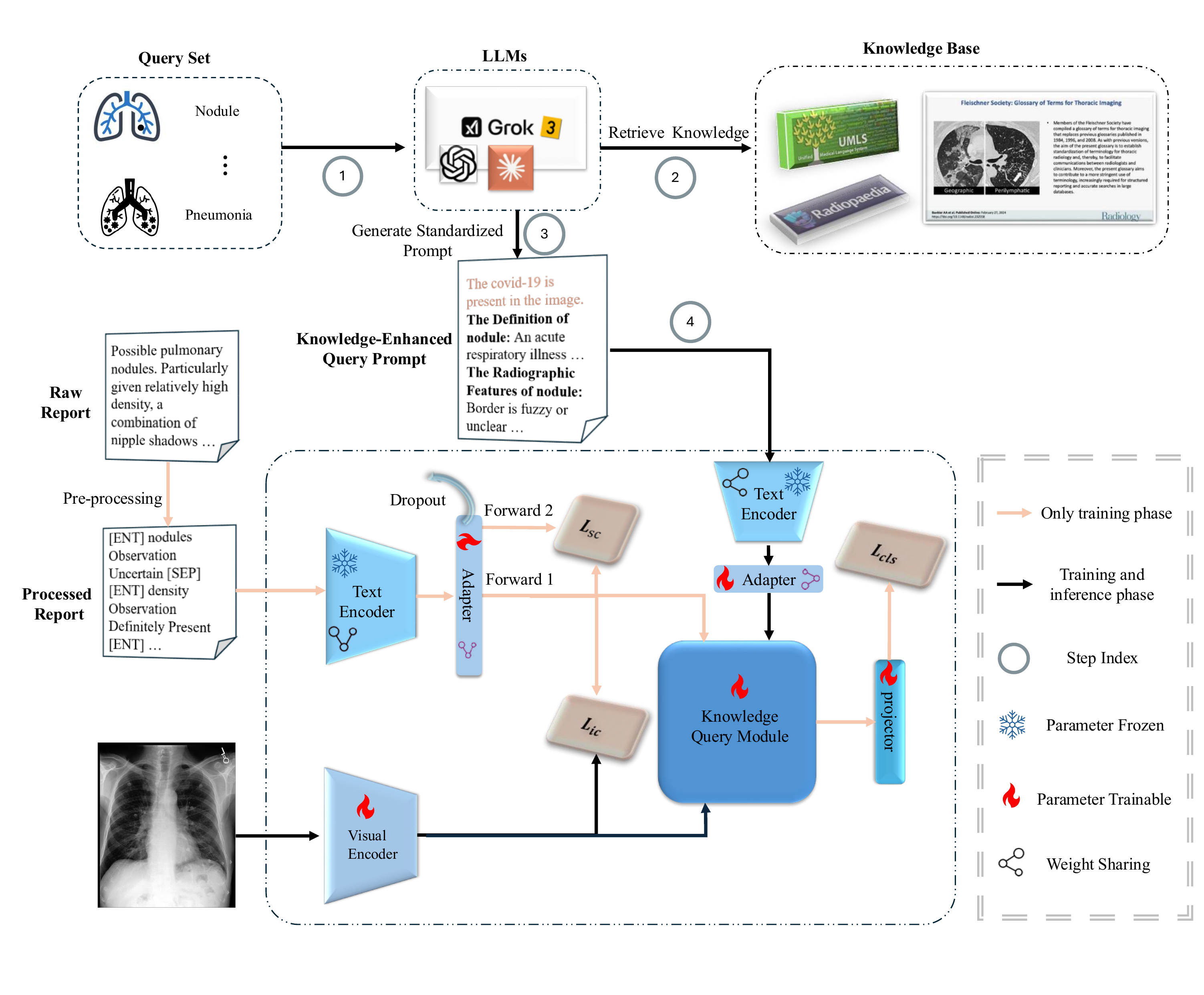}
    \caption{Overview of the proposed KEPIL framework. The model takes three inputs: a chest X-ray image, its corresponding radiology report, and a knowledge-based entity description. During training, the radiology report is parsed to extract anatomical regions, identified pathologies, and the associated confidence levels of those observations. Two forward passes with different dropout configurations are applied to the adapter to compute the semantic-aware contrastive ($\mathcal{L}_{s c}$). An instance contrastive loss ($\mathcal{L}_{ic}$) is then computed between the report text embeddings and the image embeddings. These embeddings, along with knowledge-based entity embeddings, are fed into a description query network to compute the classification loss ($\mathcal{L}_{cls}$). Knowledge enhancement is achieved by using LLMs to extract definitions and descriptions from medical knowledge bases, producing consistent and structured entity descriptions for training.}

    % includes encoding the image using a visual encoder, entity knowledge and visual descriptors with a knowledge encoder, capturing image-level entities-to-image correspondence using contrast  loss, and employing Transformer-based Fusion layers for fine-grained alignment between image patches and visual descriptor tokens using the itm and PTA losses.}
    \label{fig:framework}
\end{figure*}

Given a dataset $\mathcal{D}$ that comprises $N$ samples, each sample represented by a pair $(\mathbf{x}, \mathbf{t})$, where $\mathbf{x}$ is a radiographic image of size $H \times W$, and $\mathbf{t}$ is its corresponding radiology report. In addition, each sample is accompanied by a collection of definitions and visual depictions for $k$ medically significant entities, which we denote by $\mathbf{d} = \{d_i\}_{i=0}^k$. Our goal is to construct a vision-language model $\mathbf{\Phi}$ that, in a zero-shot scenario, computes the probability $\mathbf{\hat{y}}$ indicating the presence of a specific finding in a given test radiograph $\mathbf{x}$ with the detailed description of the finding. To achieve this, we propose a strategy comprising the following components as shown in Figure \ref{fig:framework}: (1) Collect detailed visual descriptions and general definitions of findings from the knowledge base, (2) Design prompt for pretraining and zero-shot inference, (3) Design effective network architectures and inherit knowledge from other powerful pretraining models, and (4) Devise a training protocol to align the visual and text modalities effectively.

\noindent \textbf{3.1 Knowledge Processing.}\label{sec:knowledge} Following KAD~\cite{zhang2023knowledge}, we first mitigate report noise through structured preprocessing. 
We employ RadGraph~\cite{jain2021radgraph} to extract radiological entities $\{e_i\}_{i=1}^{j}$ from a report $\mathbf{t}$ and assign each entity a semantic label $s_i \in \{\texttt{ANAT}, \texttt{OBS}\}$ with status (\texttt{Definitely Present}, \texttt{Definitely Absent}, or \texttt{Uncertain}). 
The preprocessed report is represented as $\mathbf{e} = \{e_1, s_1, \texttt{[SEP]}, \dots, e_j, s_j\}$, from which we select the top-$M$ most frequent entities to construct an entity set $\mathcal{E}$. 
To enrich textual supervision beyond standardized medical definitions (e.g., UMLS), we augment entities in $\mathcal{E}$ with structured visual descriptions by integrating radiographic features from Radiopaedia~\cite{radiopaedia} and the Fleischner glossary~\cite{hansell2008fleischner}, and employ large language models to normalize terminology, harmonize format, and complete missing attributes for semantic consistency. Based on the enriched knowledge, we construct a unified prompt template that combines a presence statement (e.g., ``\textless{}finding\textgreater{} is present in the image''), a concise definition, and characteristic radiographic features (e.g., border clarity, opacity pattern, anatomical location, fluid accumulation, and other salient signs), thereby encoding clinically grounded semantics and promoting tighter alignment between image representations and fine-grained radiological descriptors.

\noindent \textbf{3.2 Network Architecture.}
Our model adopts a dual-stream design, consisting of an image encoder and a shared text encoder for both the preprocessed report and enriched radiographic descriptions. 
Given an input chest X-ray $\mathbf{x}$, a visual backbone $\Phi_{\text{image}}$ produces patch-level features 
$v = \Phi_{\text{image}}(\mathbf{x}) \in \mathbb{R}^{P \times d}$, 
where $P$ denotes the number of patches and $d$ the embedding dimension. 
Under the ViT architecture~\cite{dosovitskiy2020image}, a learnable \texttt{CLS} token $v^{cls} \in \mathbb{R}^{d}$ summarizes the global image representation. 
For textual input, the structured report $\mathbf{e}$ and its corresponding enriched prompts $\{p_i\}_{i=0}^{k}$ are encoded via a shared text encoder $\Phi_{\text{text}}$, yielding token-level embeddings 
$t_{e^{*}} = \Phi_{\text{text}}(\mathbf{e}) \in \mathbb{R}^{T \times d}$ and 
$t_{p^{*}} = \Phi_{\text{text}}(p_i) \in \mathbb{R}^{T \times d}$, 
where $T$ is the token length. 
The encoder also produces global representations $t_{e^{*}}^{cls}$ and $t_{p^{*}}^{cls}$ to capture holistic semantics, and lightweight adapters further refine them into $t_e$ and $t_p$. To enhance fine-grained diagnostic reasoning, we introduce a Knowledge Query Module (KQM) that performs token-level cross-attention between visual patches and textual knowledge. 
Specifically, prompt embeddings $t_{p}^{cls} \in \mathbb{R}^{S \times d}$ are projected to queries $Q$, while patch-wise visual embeddings $v \in \mathbb{R}^{P \times d}$ are projected to keys $K$ and values $V$. 
The knowledge-enhanced representation is obtained via scaled dot-product attention,
$
t^{cls}_{attn} = \mathrm{softmax}\!\Bigl(\frac{QK^{\top}}{\sqrt{d}}\Bigr) V,
$
where $t^{cls}_{attn} \in \mathbb{R}^{S \times d}$. 
The fused representations are subsequently mapped to logits and transformed into diagnostic probabilities.
Figure~\ref{fig:framework} provides an overview of the overall architecture.

\noindent \textbf{3.3 Training Objectives.}\label{sec:losses}
\noindent\textbf{(\uppercase{\romannumeral 1}) \textit{Semantic-aware contrastive learning.}} To enhance the expressive power of the frozen text encoder without modifying its pretrained weights, we introduce an adapter module. This module aims to improve the discriminative capability of raw text embeddings and is trained using a contrastive loss ($\mathcal{L}_{sc}$) that maximizes consistency between the original and adapter-enhanced representations. Specifically, for a radiograph-report pair $(\mathbf{x}, \mathbf{t})$, we first extract the frozen text embedding from the \texttt{CLS} token, denoted as $t_{e^{*}}^{cls}$. The adapter then transforms this embedding as follows: $t_{e}^{cls} = \mathcal{A}(t_{e^{*}}^{cls})$,
where $\mathcal{A}$ denotes the adapter function. We repeat this process to obtain $t_{e}^{cls^{2}}$. Due to the dropout in the adapter, $t_e^{cls}$ and $t_e^{cls^{2}}$ are two perturbed views. The adapter is utilized to learn the consistency from two views of the embedding and the contrastive loss $\mathcal{L}_{sc}$ is applied to maximize the semantic consistency between $t_{e}^{cls}$ and $t_{e}^{cls^{2}}$ while ensuring they are discriminated against embeddings from other samples in a training batch $B$. Formally, the loss is defined as:
% \begin{equation}
% \mathcal{L}_{sc} (t_{e}^{cls}, t_{e}^{cls^{2}}) = -\log \frac{\exp\big(\text{s}(t_{e}^{cls}, t_{e}^{cls^{2}}) / \tau\big)}{\sum_{t_{e}^{cls^{2}} \in B} \exp\big(\text{s}(t_{e}^{cls}, t_{e}^{cls^{2}}) / \tau\big)},
% \label{eq:sc_loss}
% \end{equation}
% \begin{equation}
% \mathcal{L}_{sc}(t_{e}^{cls}, t_{e}^{cls^{2}}) = 
% - \log 
% \frac{
% \exp\left( \text{s}(t_{e}^{cls}, t_{e}^{cls^{2}}) / \tau \right)
% }{
% \sum_{t_e^{'cls} \in B \exp\left( \text{s}(t_{e}^{cls}, t_{j}^{cls^{2}}) / \tau \right)
% },
% \label{eq:sc_loss}
% \end{equation}
$ \mathcal{L}_{sc}(t_{e}^{cls},  t_{e}^{cls^{2}}) = -\log \frac{\exp\big(t_{e}^{cls},  t_{e}^{cls^{2}}) / \tau\big)}{\sum_{t_e^{'cls} \in B} \exp\big(\text{s}(t_e^{cls}, t_e^{'cls}) / \tau\big)} \label{eq:sc_loss}. $ where $\text{s}(\cdot,\cdot)$ represents the cosine similarity and $\tau$ is the temperature parameter. $B$ represents the batch.

\noindent\textbf{(\uppercase{\romannumeral 2}) \textit{Instance level alignment of radiograph-report.}} The global alignment between an image and its corresponding radiology report is accomplished using the standard image-text contrastive loss, denoted as $\mathcal{L}_{ic}$. This loss function maximizes the mutual information between the image and report representations. In our approach, we specifically utilize the report embedding because it effectively captures the overall context of the image. Given a triplet consisting of a radiograph, its report, and a description $(\mathbf{x}, \mathbf{e}, \mathbf{d})$, our goal is to enhance the similarity between the embedded image and report features, with a particular focus on their \texttt{CLS} token representations. Formally, within a training batch $B$, the loss is defined as: $ \mathcal{L}_{ic}(v^{cls},  t_e^{cls}) = -\log \frac{\exp\big(\text{s}(v^{cls}, t_e^{cls}) / \tau\big)}{\sum_{t_e^{'cls} \in B} \exp\big(\text{s}(v^{cls}, t_e^{'cls}) / \tau\big)} \label{eq:l_contrast}. $

\noindent\textbf{(\uppercase{\romannumeral 3}) \textit{\emph{Overall loss.}}}
The overall training objective integrates the contrastive report embedding loss, the instance-level image-text contrastive loss (Eq.~\ref{eq:l_contrast}) and the binary classification loss. Additionally, we employ a binary cross-entropy loss ($\mathcal{L}_{cls}$), which is applied only when queries from $\mathcal{E}$ explicitly appear or are explicitly absent in the report. If the presence of findings is uncertain, this loss is not computed. The total loss to be minimized is: $\mathcal{L} = \lambda_{1}\mathcal{L}_{cls} +  \lambda_{2}\mathcal{L}_{ic} +  \lambda_{3}\mathcal{L}_{sc}$, where $\lambda$ regulates the contribution of radiographic descriptions.

\section{Experiment Settings}
We evaluate our approach on seven publicly available chest‑X‑ray datasets for zero-shot inference:  CheXpert~\cite{irvin2019chexpert}, ChestXray‑14~\cite{wang2017chestx}, PadChest~\cite{bustos2020padchest}, RSNA Pneumonia~\cite{wu2024pneumonia}, SIIM‑ACR~\cite{siim-acr-pneumothorax-segmentation}, COVIDx CXR‑2~\cite{pavlova2022covid} and LIDC-IDRI dataset~\cite{armato2011lung}; The text encoder is initialized with BioClinicalMPBERT \cite{lai2024carzero} weights during pretraining and remains frozen. The image encoder is ViT-B/16 which utilizes M3AE for pretraining only on the MIMIC dataset \cite{lai2024carzero}. For the adapters, we use light-weighted two-layer MLPs with dropout set to 0.5 by default at the report branch. The adapter, projector, visual encoder and KQM module are trainable. The ChatGPT-4o is utilized for prompt standardization and completion as it performs the best as shown in Table \ref{tab:llms_selection}. For generating prompt variants, we include not only rephrasings but also realistic errors such as typos, omissions, and incorrect punctuation. The pretraining is conducted on two H100 GPUs.

\begin{table*}[t]
\centering
\caption{Comprehensive zero-shot evaluation on seen and unseen diseases.
Seen datasets contain diseases observed during pretraining.
Unseen settings include novel categories, rare diseases, and modality shift (CXR$\rightarrow$CT).
Best results are bold; second-best are underlined.}
\label{tab:zero_shot_all}
\resizebox{\textwidth}{!}{
\begin{tabular}{l|ccc|ccc|ccc|ccc|ccc||cc|cc|cc|c}
\toprule
\multirow{2}{*}{\textbf{Method}}
& \multicolumn{15}{c||}{\textbf{Seen Diseases (CXR)}}
& \multicolumn{7}{c}{\textbf{Unseen / Rare / Modality Shift}} \\
\cmidrule{2-23}

& \multicolumn{3}{c|}{CheXpert}
& \multicolumn{3}{c|}{ChestXray-14}
& \multicolumn{3}{c|}{PadChest-seen}
& \multicolumn{3}{c|}{RSNA}
& \multicolumn{3}{c||}{SIIM}

& \multicolumn{2}{c|}{Covid}
& \multicolumn{2}{c|}{PadChest-unseen}
& \multicolumn{2}{c|}{PadChest-rare}
& LIDC \\

& AUC & F1 & ACC
& AUC & F1 & ACC
& AUC & F1 & ACC
& AUC & F1 & ACC
& AUC & F1 & ACC

& AUC & ACC
& AUC & ACC
& AUC & ACC
& AUC \\

\midrule
ConVIRT~\cite{zhang2022contrastive}
& 52.10 & 35.61 & 57.43
& 53.15 & 12.38 & 57.88
& 63.72 & 14.56 & 73.47
& 79.21 & 55.67 & 75.08
& 64.25 & 42.87 & 53.42
& 62.78 & 63.84
& 51.17 & 61.51
& 50.37 & 60.17
& - \\

GLoRIA~\cite{huang2021gloria}
& 54.84 & 37.86 & 60.70
& 55.92 & 14.20 & 59.47
& 64.09 & 14.83 & 73.86
& 70.37 & 48.19 & 70.54
& 54.71 & 40.39 & 47.15
& 64.52 & 60.21
& 49.96 & 60.95
& 48.25 & 58.49
& - \\

BioViL~\cite{boecking2022making}
& 60.01 & 42.10 & 66.13
& 57.82 & 15.64 & 61.33
& 60.35 & 10.63 & 70.48
& 84.12 & 54.59 & 74.43
& 70.28 & 46.45 & 68.22
& 61.40 & 58.20
& 57.95 & 62.50
& 52.82 & 60.60
& - \\

BioViL-T~\cite{boecking2022making}
& 70.93 & 47.21 & 69.96
& 60.43 & 17.29 & 62.12
& 65.78 & 15.37 & 77.52
& 86.03 & 62.56 & 80.04
& 75.56 & 60.18 & 73.72
& 62.43 & 57.65
& 58.94 & 68.56
& 57.44 & 65.38
& - \\

CheXzero~\cite{tiu2022expert}
& 87.90 & 61.90 & 81.17
& 66.99 & 19.95 & 65.38
& 73.24 & 19.53 & 83.49
& 83.13 & 61.49 & 78.34
& 84.60 & 65.97 & 77.34
& 73.13 & 71.45
& 66.70 & 81.19
& 65.08 & 81.17
& - \\

MedKLIP~\cite{wu2023medklip}
& 87.97 & 63.67 & 84.32
& 72.33 & 24.18 & 79.40
& 77.87 & 26.63 & 92.44
& 86.57 & 63.28 & 79.97
& 89.79 & 72.73 & 83.99
& 76.28 & 71.96
& 60.31 & 76.69
& 59.75 & 77.84
& 45.56 \\

KAD~\cite{zhang2023knowledge}
& 89.23 & 63.25 & 86.25
& 76.49 & 29.98 & 80.49
& 80.94 & \underline{58.23} & 92.12
& 85.32 & \underline{87.09} & 81.80
& 87.39 & 68.84 & 81.73
& 74.03 & 72.42
& 73.08 & 83.23
& 72.21 & 86.54
& 57.75 \\

MAVL~\cite{phan2024decomposing}
& 90.13 & \underline{65.47} & 86.44
& 73.57 & 26.25 & \underline{82.77}
& 78.79 & 28.48 & \underline{92.56}
& \underline{86.91} & 63.41 & \underline{82.42}
& \underline{92.04} & \underline{77.95} & \underline{87.14}
& \textbf{83.86} & \underline{78.07}
& 70.42 & 84.00
& 70.06 & 84.64
& 41.81 \\

CARZero~\cite{lai2024carzero}
& \textbf{92.38} & 40.20 & \underline{86.93}
& \underline{79.64} & \underline{31.41} & 79.60
& \underline{83.97} & 25.08 & \underline{91.46}
& 80.28 & 28.28 & 41.52
& 90.90 & 75.96 & 85.03
& 75.53 & 68.55
& \underline{78.95} & \underline{85.65}
& 77.79 & \underline{91.77}
& \underline{60.57} \\

DeViDe~\cite{luo2025devide}
&  89.87 & 62.88 & \underline{87.98} & 77.61 & \underline{31.40} & 82.06 & \underline{84.26} & 58.29 & \underline{92.16} & 88.58 & 88.40 & 84.00 & 89.50 & 72.24 & 83.72 
& 73.75 & 72.34
& 75.06 & 90.84
& 73.73 & 91.09
& 47.90 \\

\midrule
\textbf{KEPIL (Ours)}
& \underline{91.21} & \textbf{65.56} & \textbf{87.88}
& \textbf{80.95} & \textbf{34.74} & \textbf{83.23}
& \textbf{85.32} & \textbf{59.79} & \textbf{93.16}
& \textbf{89.76} & \textbf{90.23} & \textbf{86.24}
& \textbf{93.02} & \textbf{79.22} & \textbf{87.46}
& \underline{79.55} & \textbf{78.12}
& \textbf{79.05} & \textbf{94.02}
& \textbf{78.75} & \textbf{95.47}
& \textbf{66.65} \\

\bottomrule
\end{tabular}}
\end{table*}

\section{Results}
\begin{figure*}[!t]
\centering

% --- 左：Figure（独立编号） ---
\begin{minipage}[t]{0.49\textwidth}
\vspace{0pt}\centering
\makeatletter\def\@captype{figure}\makeatother
\includegraphics[width=\linewidth]{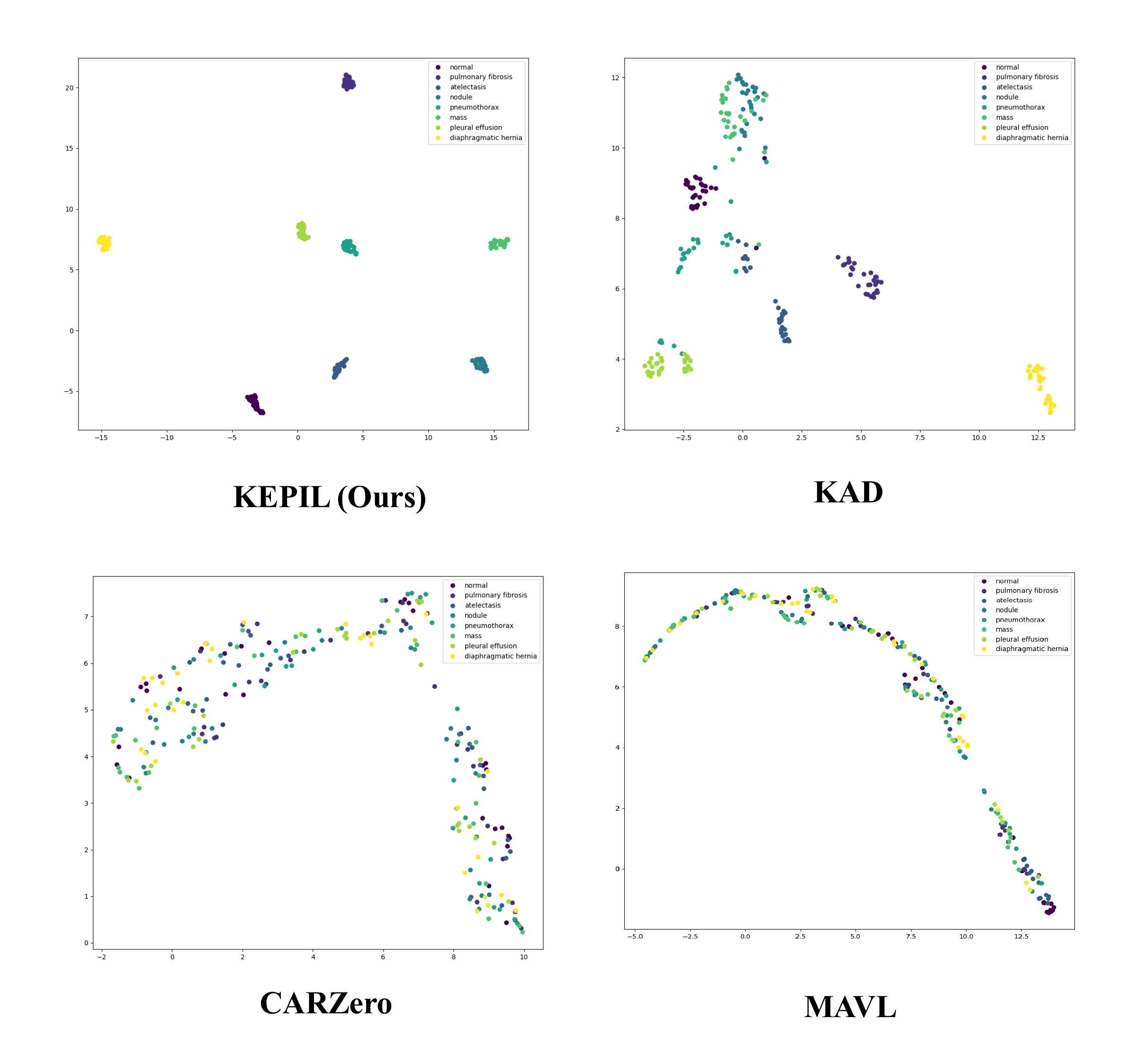}
\captionsetup{skip=4pt,justification=raggedright,singlelinecheck=false,width=\linewidth}
\caption{UMAP projection of embeddings generated by multiple clinical text encoders from 30 prompt variations per disease term. KEPIL exhibits tighter clusters, demonstrating superior robustness to prompt variations.}
\label{fig:umap}
\end{minipage}
\hfill
% --- 右：Figure（独立编号） ---
\begin{minipage}[t]{0.49\textwidth}
\vspace{0pt}\centering
\makeatletter\def\@captype{figure}\makeatother
\includegraphics[width=\linewidth]{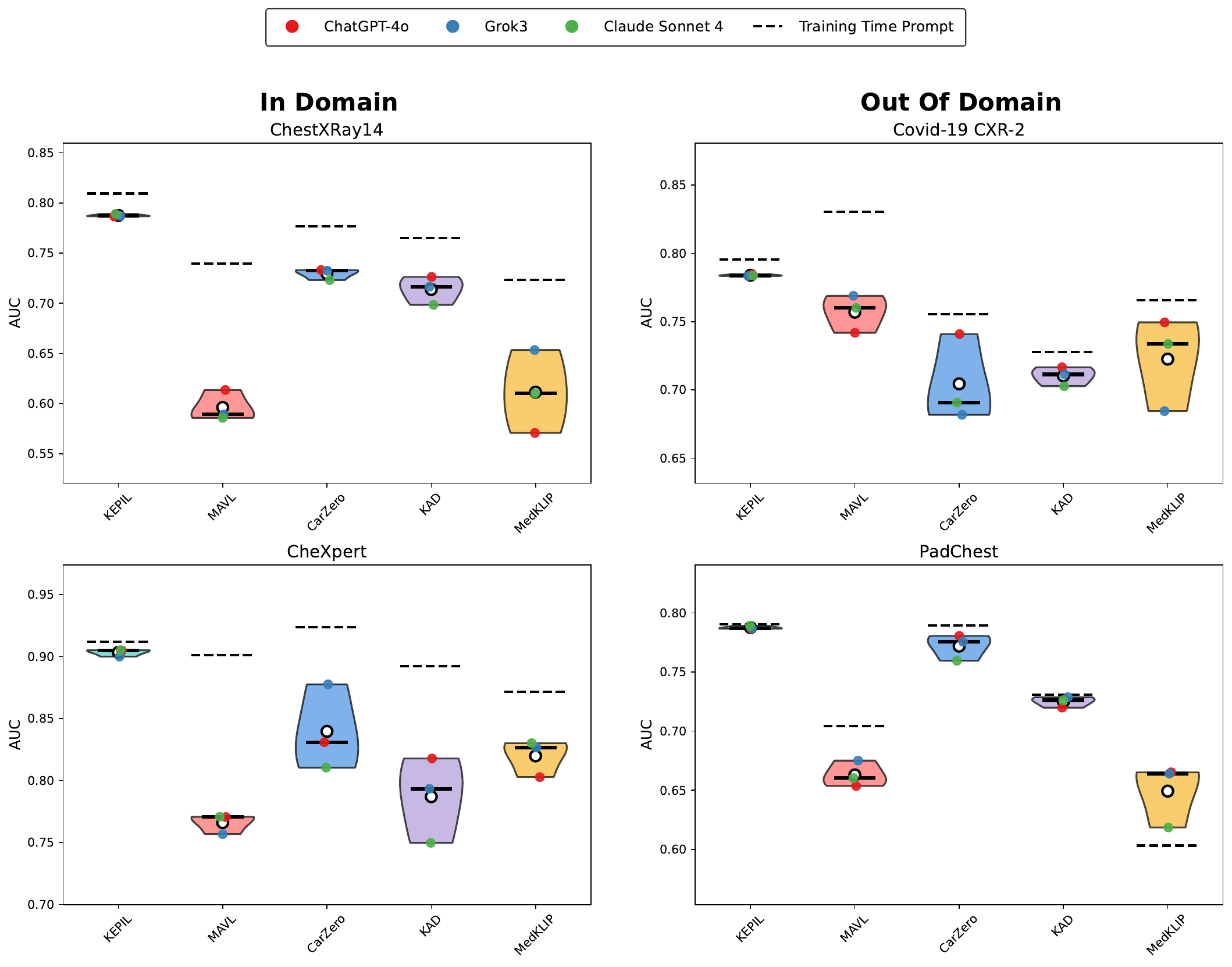}
\captionsetup{skip=4pt,justification=raggedright,singlelinecheck=false,width=\linewidth}
\caption{Evaluation of the impact of prompts from ChatGPT-4o, Grok-3, and Claude Sonnet 4 (including syntactic paraphrases, typos, omissions, punctuation variants) on chest X-ray model performance relative to the original training prompt. KEPIL demonstrates smaller performance declines.}
\label{fig:compare_robustness}
\end{minipage}

\end{figure*}

\noindent\textbf{(\uppercase{\romannumeral 1}) \textit{Classification for seen diseases.}} Table \ref{tab:zero_shot_all} presents the zero-shot classification results for \textit{seen} diseases. \textit{Seen} diseases are those that appear in the MIMIC dataset, which the model is exposed to during pretraining. Compared to other vision-language models (VLMs) pretrained on the MIMIC dataset, our proposed method, KEPIL, consistently achieves the highest performance in terms of both AUC and ACC across all five datasets. For instance, KEPIL outperforms the second-best method, CARZero, by 1.31\% on the ChestXray-14 dataset and by 2.85\% on the RSNA Pneumonia dataset in terms of AUC. Additionally, KEPIL achieves the best or second-best performance on F1 scores. These results demonstrate the model’s robustness and effectiveness in zero-shot disease detection under significant domain shifts. \\
\noindent\textbf{(\uppercase{\romannumeral 2}) \textit{Classification for unseen and rare diseases.}}
Table \ref{tab:zero_shot_all} reports zero-shot classification on diseases absent from pretraining (“unseen”) and on diseases sparsely observed during pretraining (“rare”). On the two unseen datasets, COVID-19 CXR-2 and PadChest-unseen, our proposed method, KEPIL, achieves either the best or second-best performance when compared to other vision-language models. Specifically, KEPIL demonstrates comparable performance to MAVL on the COVID-19 CXR-2 dataset and achieves notable improvements of 8.37\% in ACC on the PadChest-unseen dataset over CARZero. For rare diseases, which are included during pretraining but with very few positive sample counts, KEPIL significantly improves ACC by 3.70\% over the second-best performances. These results underscore KEPIL’s strong capability to generalize to novel or low-resource disease categories, further highlighting its potential in real-world clinical applications where data scarcity and domain shifts are common. Moreover, we further evaluated KEPIL's cross-modality generalization on CT slices from the LIDC-IDRI dataset, where it achieved a leading AUC performance that surpassed the second-best method by 6.08\%. \\% \noindent \textbf{Classification under real-world setting.} 
% Figure \ref{fig:padchest_open} illustrates the performance of our proposed method, KEPIL, on the PadChest dataset, focusing on high-impact classes with at least 50 samples. KEPIL achieves an AUC \(\geq 0.900\) for 11 conditions and \(\geq 0.700\) for 48 out of 57 conditions, demonstrating robust performance across a wide range of findings. Notably, KEPIL outperforms CheXNet (Finetuned) \cite{rajpurkar2017chexnet} on all 6 unseen diseases listed in \cite{rajpurkar2017chexnet}, including conditions such as pleural effusion, pulmonary edema, and heart insufficiency. In addition, KEPIL achieves an average AUC of 0.7851 across 193 diseases in the full Padchest dataset.
\noindent\textbf{(\uppercase{\romannumeral 3}) \textit{Robustness to Semantic Variations of Prompts.}}
Figure~\ref{fig:compare_robustness} reports zero-shot AUCs using query prompts generated by Grok3, ChatGPT-4o, and Claude Sonnet 4. Across both in-domain and out-of-domain datasets, KEPIL consistently surpasses other SOTA VLMs, demonstrating strong robustness to prompt variability—including realistic errors such as typos, omissions, and incorrect punctuation/formatting. Leveraging this robustness, KEPIL operates with minimal specification: given only a finding name, it automatically composes comprehensive, clinically grounded queries, thereby streamlining prompt creation and improving diagnostic accuracy, transparency, and efficiency. In Table~\ref{tab:kepil_combined}, we compare our $\mathcal{L}_{sc}$ with naive text augmentation. our $\mathcal{L}_{sc}$ objective stays strictly within the clinical language manifold and enforces invariance without distorting medical semantics, leading to higher robustness across all real clinical variants and better OOD performance.

% \begin{figure}[!htb]
%     \centering
%     \includegraphics[width=7.5cm]{images/Inference_type.pdf}
%     \caption{Model performance across different text input types in inference. Inputs include: (1) Finding
%     name only, (2) Name + definition, and (3) Name + definition + detailed description. Performance improves with increasingly rich textual input, highlighting the value of more informative prompts.}
%     \label{fig:factedknowledge}
% \end{figure}

% 导言区：
% \usepackage{graphicx}
% \usepackage{caption}

\begin{figure*}[!t]
\centering

% ---- 左：图（Figure 计数） ----
\begin{minipage}[t]{0.47\textwidth}
\vspace{0pt}\centering
\makeatletter\def\@captype{figure}\makeatother
\includegraphics[width=\linewidth]{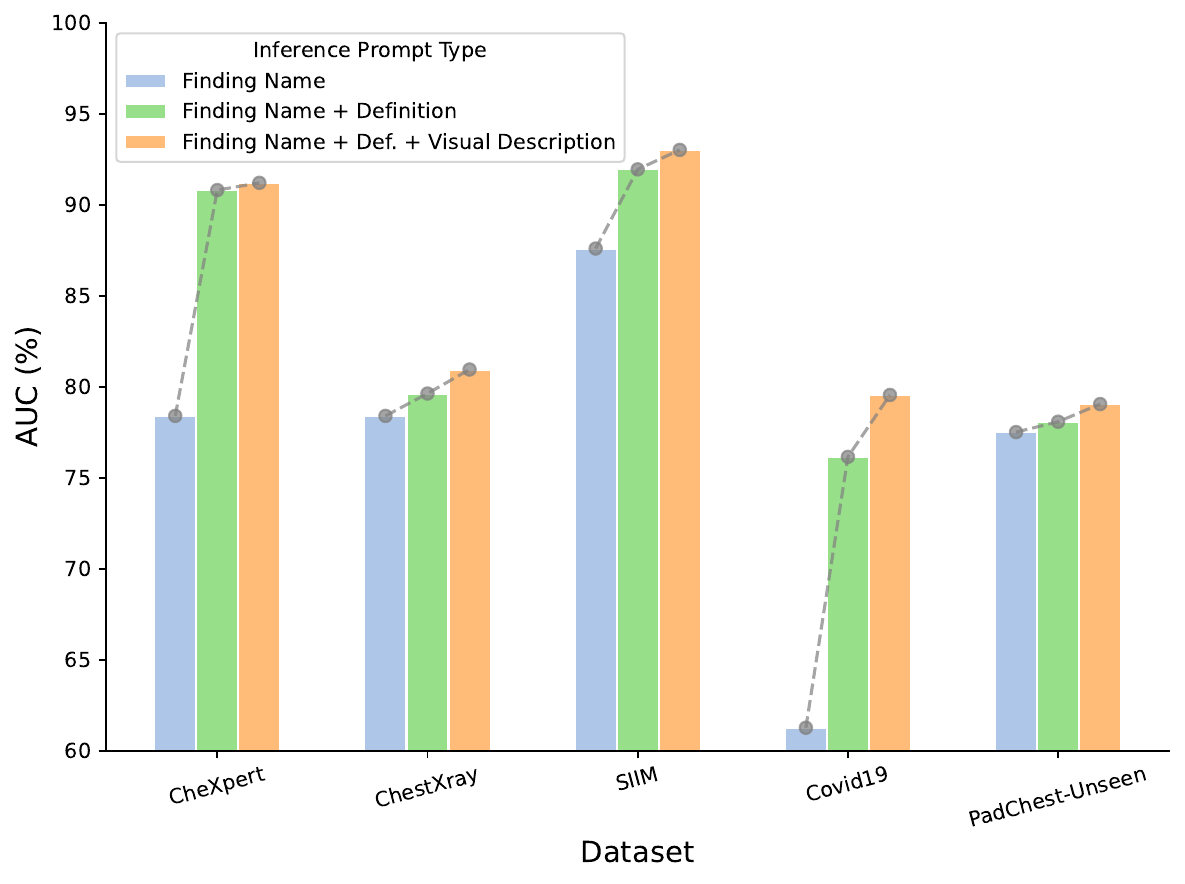}
\captionsetup{skip=4pt,justification=raggedright,singlelinecheck=false,width=\linewidth}
\caption{KEPIL's performance across incremental different inference input types. Performance improves with increasingly rich textual input, highlighting the value of more informative prompts.}
\label{fig:factedknowledge}

\end{minipage}
\hfill
% ---- 右：表（Table 计数；caption 在上方） ----
\begin{minipage}[t]{0.49\textwidth}
\vspace{0pt}
\captionsetup{skip=4pt,justification=raggedright,singlelinecheck=false,width=\linewidth}
\captionof{table}{Ablation study of the effects of enriched prompts, the inclusion and placement of \(\mathcal{L}_{sc}\), and dropout rate on zero-shot inference performance.}
\label{table:ablations_combined}
\vspace{2pt}
\setlength{\tabcolsep}{6pt}\renewcommand{\arraystretch}{1.1}
\resizebox{\linewidth}{!}{
\begin{tabular}{lcc}
\hline
\multicolumn{1}{c}{\textbf{Ablation}} & \textbf{ChestXRay14} & \textbf{CheXpert} \\
\hline
\multicolumn{3}{l}{\emph{Effect of Enriched Prompt and \(\mathcal{L}_{sc}\)}}\\
Base (\(\mathcal{L}_{cls}\) + \(\mathcal{L}_{ic}\)) & 76.89 & 87.07 \\
+ Enriched Prompt (EP)                               & 79.45 & 90.47 \\
+ EP + \(\mathcal{L}_{sc}\)                         & \textbf{80.95} & \textbf{91.21} \\
\hline
\multicolumn{3}{l}{\emph{Effect of Dropout Rate in \(\mathcal{L}_{sc}\)}}\\
Dropout = 0.3 & 80.11 & 90.98 \\
Dropout = 0.4 & 80.29 & 91.41 \\
Dropout = 0.5 & \textbf{80.95} & \textbf{91.21} \\
Dropout = 0.6 & 80.24 & 91.24 \\
\hline
\multicolumn{3}{l}{\emph{Effect of the Position of \(\mathcal{L}_{sc}\)}}\\
Report branch & \textbf{80.95} & \textbf{91.21} \\
Prompt branch & 79.72 & 90.52 \\
Report branch + Prompt branch & 80.19 & 91.15 \\
\hline
\end{tabular}}

\vspace{10pt} % 控制两个表之间间距

% ==========================
% 追加的 LLM selection 表
% ==========================

\captionsetup{skip=4pt,justification=raggedright,singlelinecheck=false,width=\linewidth}
\captionof{table}{Evaluation of LLM generated descriptions against ground-truth.}
\label{tab:llms_selection}
\vspace{2pt}
\setlength{\tabcolsep}{5pt}\renewcommand{\arraystretch}{1.1}
\resizebox{\linewidth}{!}{
\begin{tabular}{lcccc}
\hline
Metric & ChatGPT-4o & Mixtral-8$\times$7B & Medorca-2x7B & PMC-llama-7B \\
\hline
BLEU-4 & \textbf{0.1180} & 0.1030 & 0.0018 & 0.0033 \\
ROUGE-L & \textbf{0.2695} & 0.2328 & 0.1461 & 0.1460 \\
F1 RadGraph & \textbf{0.2384} & 0.2046 & 0.0639 & 0.0979 \\
\hline
\end{tabular}}

\end{minipage}

\end{figure*}

\noindent(\textbf{{\romannumeral 4}) \textit{Richer Textual Inputs Enhance Diagnostic Performance.}}
\noindent We investigated the impact of varying levels of textual detail on KEPIL's performance. As shown in Figure~\ref{fig:factedknowledge}, model performance consistently improves with richer textual inputs. Specifically, using only finding names (e.g., “Atelectasis”) results in the lowest AUC scores, while supplementing the names with definitions from the UMLS knowledge base leads to notable gains. The highest performance is achieved when definitions are further enriched with detailed descriptions, highlighting the model’s ability to establish a connection between diagnosis and finding-related features. This finding underscores the importance of incorporating rich and structured textual context to enhance the model's robustness and classification accuracy.
\\
\noindent(\textbf{{\romannumeral 5}) \textit{Enriched Prompt and $\mathcal{L}_{sc}$ trigger Significant Performance Gains.}}
% We ablated the impact of enriched query prompts, $\mathcal{L}_{sc}$ and $\mathcal{L}_{itm}$. As shown in Table~\ref{table:ablations1}, removing these components led to over a 4\% drop in AUC on two datasets. 
We conduct an ablation study to assess the contribution of each component in the KEPIL framework by progressively adding the Enriched Prompt (EP) and $\mathcal{L}_{sc}$ to the base optimization strategy ($\mathcal{L}_{cls} + \mathcal{L}_{ic}$). As shown in Table~\ref{table:ablations_combined}, incorporating enriched prompts significantly improves zero-shot classification performance, boosting the AUC from 76.89\% to 79.45\% on ChestXray14 and from 87.07\% to 90.47\% on CheXpert. Adding $\mathcal{L}_{sc}$ with a dropout rate of 0.5 further enhances performance, particularly on ChestXray14, reaching the highest AUC of 80.95\%.   Experiments with dropout rates between 0.3 and 0.6 show 0.5 to be optimal. We also evaluated where to apply \(\mathcal{L}_{sc}\) and found that placing it in the report branch yields the greatest improvement.  Overall, the complete KEPIL model achieves a performance gain of over 4\% in AUC compared to the base setup across datasets, demonstrating the effectiveness of enriched prompts and the proposed \(\mathcal{L}_{sc}\).

\begin{table*}[t]
\centering
\caption{Comparison of \(\mathcal{L}_{sc}\) and text side augmentation's impact on zero-shot inference. \(\mathcal{L}_{sc}\) remains strictly within the clinical language manifold, yielding real-world robustness. Achiving at most 23.96\% improvement compare to CARZero. Numbers in parentheses indicate absolute improvement over CARZero.}
\label{tab:kepil_combined}
\resizebox{\textwidth}{!}{
\begin{tabular}{lcc|cc|cccc}
\toprule
\multirow{2}{*}{\textit{Pretraining Setting}} 
& \multicolumn{2}{c|}{\textit{In-domain}} 
& \multicolumn{2}{c|}{\textit{OOD}} 
& \multicolumn{4}{c}{\textit{Prompt Variants (on CheXpert)}} \\

& CheXpert & CXR14 
& PadChest & COVID 
& Abbreviation & Multilingual & Clinician Jargon & Medical Synonym \\

\midrule
CARZero
& - & \- 
& - & -
&78.81	&62.07	&79.85	&85.21 \\

KEPIL+ \(\mathcal{L}_{sc}\) 
& 91.21 & \textbf{80.95} 
& \textbf{79.05} & \textbf{79.55} 
& 89.54 ($\uparrow$10.73) & 86.03 ($\uparrow$23.96) & \textbf{89.89 ($\uparrow$10.04)} & \textbf{90.62 ($\uparrow$5.41)}  \\

+ Text augmentation 
& \textbf{91.78} & 80.87 
& 77.28 & 75.57 
& \textbf{89.63} & \textbf{87.84} & 83.15 & 89.87 \\

\bottomrule
\end{tabular}
}
\end{table*}

\section{Conclusion}
We introduce KEPIL, a robust knowledge-enhanced vision-language framework designed to address the critical limitations of prompt sensitivity and poor generalization in open-vocabulary disease detection for radiological imaging. Unlike prior VLMs that rely on fixed or template-based prompts, KEPIL integrates medical ontologies, radiographic descriptors, and entity-level representations into the vision-language pipeline. KEPIL establishes a semantically aligned and clinically grounded interface between text and image modalities with a semantic-aware contrastive learning objective and a prompt standardization protocol, leading to enhanced stability under prompt perturbations and improved generalization to rare and unseen diseases.

\bibliographystyle{splncs04}
\bibliography{reference}

% \appendix
% \section{Appendix}
% You may include other additional sections here.

\end{document}